\documentclass[conference]{IEEEtran}
\usepackage[T1]{fontenc}
\usepackage[utf8]{inputenc}
\usepackage{amsmath,amssymb,amsfonts}
\usepackage{graphicx}
\usepackage{textcomp}
\usepackage{xcolor}
\usepackage{colortbl}
\usepackage{booktabs}
\usepackage{multirow}
\usepackage{array}
\usepackage{url}
\usepackage{pgfplots}
\usepackage{pgfplotstable}
\usepackage{tikz}
\usetikzlibrary{arrows.meta,positioning,shapes.misc,fit,backgrounds,calc,patterns}
\pgfplotsset{compat=1.18}

\def\BibTeX{{\rm B\kern-.05em{\sc i\kern-.025em b}\kern-.08em
    T\kern-.1667em\lower.7ex\hbox{E}\kern-.125emX}}

\newcommand{\ugm}{$\mu$g/m$^3$}
\newcommand{\PM}{PM$_{2.5}$}

\begin{document}

\title{Conformal \PM{} Mapping Under Spatial Covariate Shift:
Satellite--Reanalysis Fusion for Africa's Green Industrial Transition}

\author{%
\IEEEauthorblockN{Yaw Osei Adjei}
\IEEEauthorblockA{Dept.\ of Computer Science\\
Kwame Nkrumah Univ.\ of Science and Technology\\
Kumasi, Ghana\\
yoadjei@st.knust.edu.gh}
\and
\IEEEauthorblockN{Davis Opoku}
\IEEEauthorblockA{Dept.\ of Computer Science\\
Kwame Nkrumah Univ.\ of Science and Technology\\
Kumasi, Ghana\\
dopoku94@st.knust.edu.gh}
\and
\IEEEauthorblockN{Ephraim Abotsi}
\IEEEauthorblockA{Dept.\ of Computer Science\\
Kwame Nkrumah Univ.\ of Science and Technology\\
Kumasi, Ghana\\
eabotsi4@st.knust.edu.gh}
\and
\IEEEauthorblockN{Kwadwo Owusu Amanqua}
\IEEEauthorblockA{Dept.\ of Computer Science\\
Kwame Nkrumah Univ.\ of Science and Technology\\
Kumasi, Ghana\\
kamanquaowusu@st.knust.edu.gh}
\and
\IEEEauthorblockN{Oliver Kornyo}
\IEEEauthorblockA{Dept.\ of Computer Science\\
Kwame Nkrumah Univ.\ of Science and Technology\\
Kumasi, Ghana\\
oliverkornyo@knust.edu.gh}
\and
\IEEEauthorblockN{Elisha Soglo-Ahianyo}
\IEEEauthorblockA{Dept.\ of Computer Science\\
Kwame Nkrumah Univ.\ of Science and Technology\\
Kumasi, Ghana\\
esogloahianyo2@st.knust.edu.gh}
\and
\IEEEauthorblockN{Cephas Anertey Abbey}
\IEEEauthorblockA{Dept.\ of Computer Science\\
Kwame Nkrumah Univ.\ of Science and Technology\\
Kumasi, Ghana\\
caabbey@st.knust.edu.gh}
\and
\IEEEauthorblockN{Ransford Oppong}
\IEEEauthorblockA{Dept.\ of Computer Engineering\\
Kwame Nkrumah Univ.\ of Science and Technology\\
Kumasi, Ghana\\
oppong.rans@gmail.com}}

\maketitle

\begin{abstract}
Africa's green industrialization imperative demands transparent, trustworthy air
quality monitoring infrastructure.  We present a satellite--reanalysis \PM{}
fusion system trained on 2{,}068{,}901 records from 404 monitoring locations
in 29 African countries (OpenAQ, 2017--2022), combining LightGBM with
rigorous leakage-resistant spatial cross-validation and conformal prediction
to quantify both predictions and their geographic applicability limits.
Under leakage-resistant 5-fold location-grouped spatial cross-validation---the
evaluation protocol required for policy deployment---LightGBM achieves
RMSE\,$=30.83\pm5.07$\,\ugm, MAE\,$=14.54\pm1.66$\,\ugm, R$^2=0.134\pm0.023$,
and macro F1\,$=0.336\pm0.018$ over six AQI bins.
This R$^2$ is substantially below random-split benchmarks ($>$0.90) but
reflects true geographic generalisation difficulty rather than model failure:
it tells decision-makers exactly where and how much to trust the predictions.
Split conformal prediction targeting 90\,\% marginal coverage reveals severe
East Africa degradation (actual PICP\,=\,65.3\,\% vs.\ nominal 90\,\%),
consistent with medium-strength covariate shift (humidity KS\,$=0.2237$,
sat\_pblh KS\,$=0.2558$); in operational terms, this means \textsc{Unreliable}
flags must block climate finance disbursement in that region until additional
monitors are deployed.
We operationalise these findings through (1) regional reliability flags
(High/Medium/Low/Unreliable) keyed to quantified validation metrics and
(2) a monitor prioritisation score directing infrastructure expansion toward
highest-burden unmonitored populations.
These tools directly support Africa's green industrial transition, aligning
with SDGs 3.9, 7.1.2, 9, 11.6.2, and 13, and providing measurement
infrastructure for climate finance verification and environmental justice
for the 405 million extremely poor Sub-Saharan Africans exposed to unsafe
air quality~\cite{rentschler2023}.
\end{abstract}


\section{Introduction}

Africa faces an industrial paradox: it must industrialise to meet development
goals yet cannot replicate the polluting pathways of the Global North.
Air pollution kills an estimated 155\,per\,100{,}000 people in Africa---nearly
double the global average of 85.6\,per\,100{,}000~\cite{hei2024}.
Five of the ten most polluted countries globally are in Africa~\cite{iqair2025};
Chad alone averages 91.8\,\ugm{} \PM{}---18\,$\times$ the WHO guideline of
5\,\ugm~\cite{iqair2025}.
In 2023 ambient air pollution caused 4.9 million deaths globally, a 24\,\%
increase from 2013~\cite{hei2024}.
The Global Burden of Disease (GBD) 2021 study attributes 7.83 million deaths
and 231.51 million DALYs to \PM{} exposure~\cite{gbd2021rf}, with Sub-Saharan
Africa (SSA) bearing the highest household air pollution DALY rate globally at
4{,}044\,per\,100{,}000~\cite{gbd2021hap}.
In 2019 alone, 383{,}419 deaths from ambient air pollution occurred across
Africa~\cite{unep2023}, while African children lose an estimated 1.96 billion
IQ points annually from \PM{} exposure~\cite{landrigan2021}.

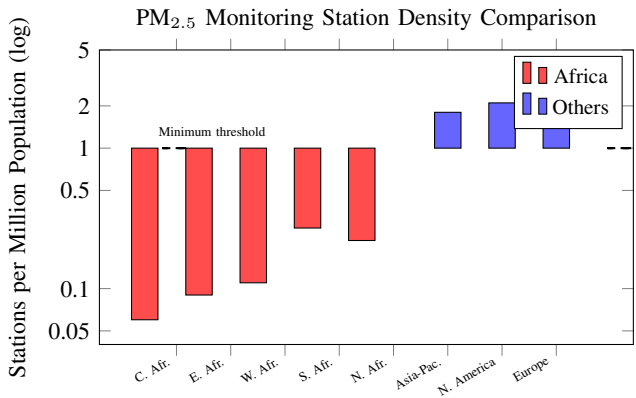
\begin{figure}[!t]
\centering
\begin{tikzpicture}
\begin{axis}[
  ybar, ymode=log,
  ymin=0.04, ymax=5,
  width=0.98\columnwidth, height=5.5cm,
  ylabel={Stations per Million Population (log)},
  xtick={1,2,3,4,5,6,7,8},
  xticklabels={C.~Afr.,E.~Afr.,W.~Afr.,S.~Afr.,
               N.~Afr.,Asia-Pac.,N.~America,Europe},
  x tick label style={font=\tiny, rotate=30, anchor=east},
  ytick={0.05,0.1,0.5,1,2,5},
  yticklabels={0.05,0.1,0.5,1,2,5},
  tick label style={font=\small},
  label style={font=\small},
  bar width=10pt,
  title={\small \PM{} Monitoring Station Density Comparison},
  title style={yshift=-2pt},
]
\addplot[fill=red!70] coordinates
  {(1,0.06)(2,0.09)(3,0.11)(4,0.27)(5,0.22)};
\addplot[fill=blue!60] coordinates
  {(6,1.8)(7,2.1)(8,2.9)};
\addplot[mark=none, black, dashed, thick]
  coordinates {(0.4,1.0)(8.6,1.0)};
\node[font=\tiny, anchor=south west] at (axis cs:0.5,1.05)
  {Minimum threshold};
\legend{\footnotesize Africa, \footnotesize Others}
\end{axis}
\end{tikzpicture}
\caption{PM$_{2.5}$ monitoring station density (stations per million
  population, log scale).  African sub-regions are 10--50$\times$ more
  sparse than Europe and North America.  Central and East Africa ($<0.1$
  stations/million) represent the most acute monitoring deserts.
  Dashed line indicates an aspirational minimum of 1 station per million.
  Sources: IQAir 2025~\cite{iqair2025}; Think Global Health 2024~\cite{tgh2024};
  WHO v6.1 2024~\cite{who2024}.}
\label{fig:density}
\end{figure}

The continent faces a simultaneous monitoring crisis.
Only one public real-time air quality monitoring station exists per 3.7~million
people in Africa, compared with one per 500{,}000 in the USA and
Europe~\cite{iqair2025,tgh2024} (Fig.~\ref{fig:density}).

The WHO Ambient Air Quality Database v6.1 covers only 41 cities across 11 of
47 SSA nations~\cite{who2024}.
Zero African countries meet the WHO annual \PM{} standard, and only 17 of 54
African countries have national air quality standards in law~\cite{unep2021}.

These gaps carry severe economic consequences.
Air pollution costs an average of 6.5\,\% of GDP across Africa~\cite{cleanair2023},
and projected health and productivity losses in six major African cities
(Accra, Cairo, Johannesburg, Lagos, Nairobi, Yaound\'e) will exceed
US\$138~billion by 2040~\cite{cleanair2023dalberg}.
Yet SSA received a 91\,\% decline in outdoor air quality funding in 2023 and
now receives less than 1\,\% of global clean air investment~\cite{cleanair2025}.
Only US\$43.7~billion of Africa's estimated US\$189~billion annual climate
finance need was mobilised in 2021--2022~\cite{cpi2024}.

Satellite and reanalysis products offer a monitoring path, but only 22.5\,\%
of Earth observation datasets incorporate any uncertainty quantification~\cite{singh2024},
and AI models applied uncritically in the Global South risk encoding
overconfident predictions into policy-critical decisions~\cite{mcgovern2022}.
The challenge is therefore not data availability but \emph{trustworthy inference}:
models must quantify their own geographic applicability limits alongside their
predictions.

We contribute three concrete tools toward this vision:
\begin{enumerate}
  \item \textbf{Leakage-resistant spatial validation.}  Location-grouped
    cross-validation~\cite{roberts2017} prevents temporal autocorrelation
    leakage and reveals a true spatial generalisation gap
    (R$^2=0.134$ vs.\ $>$0.90 in random-split benchmarks~\cite{westervelt2025}).
  \item \textbf{Regional conformal prediction with explicit failure detection.}
    Split conformal prediction~\cite{shafer2008,romano2019} targets
    90\,\% marginal coverage but exposes severe East Africa failure
    (65.3\,\% actual) via Kolmogorov--Smirnov covariate shift diagnostics.
    A 24.7~percentage-point coverage shortfall means predictions in that
    region cannot be used to justify climate finance disbursement or
    industrial permitting without additional ground monitoring.
  \item \textbf{Operational reliability and prioritisation framework.}
    Regional reliability flags and a monitor prioritisation score
    (Eq.~\ref{eq:priority}) guide policy deployment and infrastructure
    investment toward highest-burden unmonitored populations.
\end{enumerate}

\section{Related Work}

\textbf{\PM{} Mapping in Africa.}
Westervelt et al.~\cite{westervelt2025} achieved R$^2=0.91$
and MAE\,$=9.1$\,\ugm{} for daily West African \PM{} at 1\,km$^2$ resolution
using XGBoost---under random cross-validation.
Zhang et al.~\cite{zhang2021} demonstrated R$^2=0.80$ and
RMSE\,$=9.40$\,\ugm{} for South Africa via random forest with random CV.
Bai et al.~\cite{bai2023} synthesised 833 \PM{} publications, identifying
tree-based ensembles as globally dominant.
None of these benchmarks enforces strict spatial cross-validation,
masking the field-scale generalisation gap critical for policy deployment.

\textbf{Covariate Shift and Transfer Learning.}
Tibshirani et al.~\cite{tibshirani2019} proved that standard conformal
prediction's coverage guarantee collapses under covariate shift, with actual
coverage degrading to 82.2\,\% of a nominal 90\,\%; weighted conformal
prediction restores it.
Yadav et al.~\cite{yadav2024} and Gupta et al.~\cite{gupta2024} address
domain adaptation for cross-regional \PM{} estimation.
Pournaderi et al.~\cite{pournaderi2024} model training-conditional shift in
high dimensions; Yang et al.~\cite{yang2024} develop doubly robust calibration
under observational covariate shift.

\textbf{Spatial Conformal Prediction.}
Mao et al.~\cite{mao2024} establish valid model-free spatial prediction via
approximate exchangeability under infill asymptotics.
Angelopoulos \& Bates~\cite{angelopoulos2023} provide the practitioner guide.
Kang et al.~\cite{kang2025} introduce GeoConformal Prediction, integrating
Tobler's law (geographic weighting) into conformal nonconformity scores.

\textbf{Area of Applicability.}
Meyer \& Pebesma~\cite{meyer2021} formalise the Area of Applicability (AOA) to
flag predictions outside the training covariate distribution.
Mil\`{a} et al.~\cite{mila2022} develop nearest-neighbour dissimilarity
matching (NNDM) cross-validation to avoid optimistic performance estimation.

\textbf{Feature Attribution in Air Quality.}
Houdou \& Alyousifi~\cite{houdou2024} survey SHAP and LIME for \PM{} models.
Just et al.~\cite{just2020} applied TreeSHAP to XGBoost \PM{} models in the
USA, finding spatial CV RMSE 48\,\% higher than random CV RMSE---directly
motivating our evaluation protocol.

\textbf{Trustworthy AI in the Global South.}
McGovern et al.~\cite{mcgovern2022} argue that environmental AI must
explicitly quantify uncertainty and applicability limits.
Singh et al.~\cite{singh2024} find only 22.5\,\% of Earth observation datasets
include uncertainty estimates, underscoring a widespread practice gap.

\section{Data and Feature Construction}

\subsection{Ground Truth}

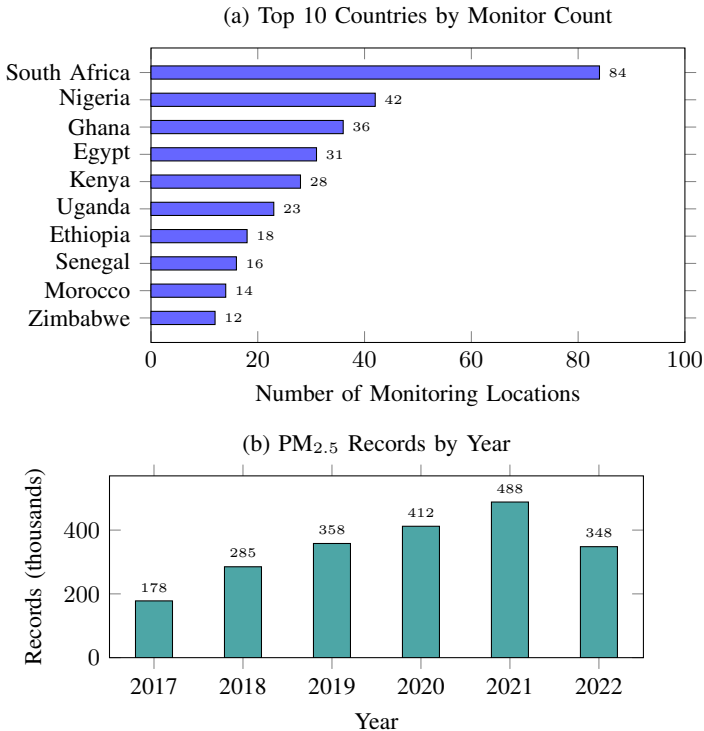
\begin{figure}[!t]
\centering
\begin{tikzpicture}
\begin{axis}[
  xbar, xmin=0, xmax=100,
  width=0.98\columnwidth, height=5.5cm,
  xlabel={Number of Monitoring Locations},
  ytick=data,
  yticklabels={South Africa,Nigeria,Ghana,Egypt,
               Kenya,Uganda,Ethiopia,Senegal,Morocco,Zimbabwe},
  bar width=5pt,
  nodes near coords, nodes near coords align={horizontal},
  every node near coord/.append style={font=\tiny},
  tick label style={font=\small},
  label style={font=\small},
  title={\small (a) Top 10 Countries by Monitor Count},
  title style={yshift=-2pt},
]
\addplot[fill=blue!60] coordinates {
  (84,9)(42,8)(36,7)(31,6)(28,5)
  (23,4)(18,3)(16,2)(14,1)(12,0)};
\end{axis}
\end{tikzpicture}

\vspace{4pt}

\begin{tikzpicture}
\begin{axis}[
  ybar, ymin=0, ymax=570,
  width=0.98\columnwidth, height=4cm,
  xlabel={Year},
  ylabel={Records (thousands)},
  xtick={1,2,3,4,5,6},
  xticklabels={2017,2018,2019,2020,2021,2022},
  bar width=14pt,
  nodes near coords, nodes near coords align={vertical},
  every node near coord/.append style={font=\tiny},
  tick label style={font=\small},
  label style={font=\small},
  title={\small (b) \PM{} Records by Year},
  title style={yshift=-2pt},
]
\addplot[fill=teal!70] coordinates {
  (1,178)(2,285)(3,358)(4,412)(5,488)(6,348)};
\end{axis}
\end{tikzpicture}
\caption{Data overview. (a)~Top 10 African countries by monitoring location
  count (total 404 locations, 29 countries); (b)~\PM{} records by year
  (total 2{,}068{,}901 records, 2017--2022, OpenAQ~\cite{openaq}).}
\label{fig:data}
\end{figure}

Real-time \PM{} observations were aggregated from OpenAQ~\cite{openaq},
a crowd-sourced air quality platform.
A uniform audit removed records with missing critical fields and filtered
values outside the physical range $0 < \text{PM}_{2.5} < 1000$\,\ugm.
The cleaned dataset comprises \textbf{2{,}068{,}901 observations from 404
monitoring locations spanning 29 African countries} (2017--2022).
Fig.~\ref{fig:data} summarises geographic and temporal coverage.

\subsection{Predictor Variables}

Table~\ref{tab:features} lists the five feature groups.
Satellite aerosol products follow the MODIS Collection~6 MAIAC
algorithm~\cite{lyapustin2018}.
Meteorological covariates are from ERA5 reanalysis~\cite{hersbach2020}.
Planetary boundary layer height (PBLH) is from MERRA-2~\cite{merra2}.
Population density is from WorldPop 2020~\cite{tatem2017}.
All predictors are regridded to 0.1$^\circ$ resolution and
standardised within each location-year group to mitigate temporal drift.

\begin{table}[!t]
\centering
\caption{Feature Groups Used in the Final Model}
\label{tab:features}
\setlength\tabcolsep{4pt}
\begin{tabular}{ll}
\toprule
\textbf{Group} & \textbf{Variables} \\
\midrule
Geographic    & latitude, longitude \\
Temporal      & month$_{\sin}$, month$_{\cos}$, harmattan\_flag \\
Atmospheric   & sat\_aot, sat\_no2, sat\_pblh \\
Meteorological & temperature, humidity, pressure, \\
              & wind\_speed, precipitation, clouds \\
Demographic   & population density \\
\bottomrule
\end{tabular}
\end{table}

\subsection{Target: AQI Classification}

\PM{} is binned into six AQI-style categories (Table~\ref{tab:aqi}) to
support both regression evaluation (RMSE, MAE, R$^2$) and classification
assessment (accuracy, macro F1).

\begin{table}[!t]
\centering
\caption{AQI-Style \PM{} Bins}
\label{tab:aqi}
\small
\setlength\tabcolsep{3pt}
\begin{tabular}{p{2.6cm}p{1.4cm}p{2.8cm}}
\toprule
\textbf{Category} & \textbf{Range (\ugm)} & \textbf{Health Implication} \\
\midrule
Good                     & [0,\,12]        & Negligible risk \\
Moderate                 & (12,\,35]       & Sensitive groups \\
Unhealthy for Sensitive (USG) & (35,\,55] & Sensitive groups at risk \\
Unhealthy                & (55,\,150]      & General population at risk \\
Very Unhealthy           & (150,\,250]     & Widespread health risk \\
Hazardous                & $(250,\infty)$  & Emergency conditions \\
\bottomrule
\end{tabular}
\end{table}

\section{Methods}

\subsection{Spatial Cross-Validation}

To prevent train-test leakage from spatial autocorrelation, we employ
5-fold location-grouped cross-validation~\cite{roberts2017}: all
observations from a monitoring location are assigned to one fold, ensuring
that test folds contain completely held-out stations.
Folds are stratified by sub-region (North, West, Central, East, Southern
Africa) to balance covariate shift assessment.

\subsection{Models and Baselines}

Four models are evaluated: (1)~\textbf{Seasonal Naive}---month-wise mean
\PM{} at each location; (2)~\textbf{Ridge Regression} ($\alpha=1.0$);
(3)~\textbf{LightGBM}~\cite{ke2017}---200 estimators, max depth~8,
learning rate~0.1, feature fraction~0.9;
(4)~\textbf{XGBoost}~\cite{chen2016}---comparable hyperparameters.
Hyperparameters are tuned on 20\,\% of training data via 3-fold random CV
to avoid spatial leakage.

\subsection{Conformal Prediction}

Split conformal prediction~\cite{shafer2008} targeting $(1-\alpha)=90\,\%$
nominal coverage proceeds as:
\begin{enumerate}
  \item Split training data into a proper training set (80\,\%) and
        calibration set (20\,\%).
  \item Train LightGBM on the proper training set; generate point
        predictions $\hat{y}_i$ on the calibration set.
  \item Compute nonconformity scores: $R_i = |y_i - \hat{y}_i|$.
  \item For each test sample, set $\hat{q} = $ the
        $\lceil(n{+}1)(1{-}\alpha)/n\rceil$-th order statistic of $\{R_i\}$
        and return interval $[\hat{y}\,{-}\,\hat{q},\; \hat{y}\,{+}\,\hat{q}]$.
\end{enumerate}
This provides a marginal coverage guarantee under exchangeability~\cite{shafer2008}:
$\Pr(y \in \text{interval}) \geq 1-\alpha$.
\emph{Conditional} coverage under covariate shift is not guaranteed and must
be verified regionally~\cite{tibshirani2019}.

\subsection{Covariate Shift Diagnostics}

For each feature $f$ we compute the Kolmogorov--Smirnov statistic between
training and East African test distributions:
\[
  \text{KS}(f) = \sup_x \bigl|F_{\text{train}}^f(x) - F_{\text{test}}^f(x)\bigr|.
\]
Severity is Low ($<0.10$), Medium ($0.10$--$0.25$), or High ($>0.25$).

\subsection{Regional Reliability Flags}
\label{sec:reliability}

For each sub-region $r$ with regional regression score $R^2_r$ and mean
conformal interval half-width $w_r$, we assign:
\begin{equation}
\text{Reliability}(r) =
\begin{cases}
\textsc{High}       & R^2_r > 0.1 \;\wedge\; w_r < 40 \\
\textsc{Medium}     & R^2_r > 0   \;\wedge\; w_r < 80 \\
\textsc{Low}        & R^2_r > -0.5 \;\wedge\; w_r < 100 \\
\textsc{Unreliable} & \text{otherwise.}
\end{cases}
\label{eq:reliability}
\end{equation}

\subsection{Monitor Prioritisation Score}

To target new station deployment toward highest-burden unmonitored areas:
\begin{equation}
  \text{priority}(x) = w(x) \cdot \log\!\bigl(1 + \rho_{\text{pop}}(x)\bigr),
  \label{eq:priority}
\end{equation}
where $w(x)$ is the conformal interval half-width at location $x$ (proxy for
model uncertainty) and $\rho_{\text{pop}}(x)$ is population density in
persons\,km$^{-2}$.

\subsection{Feature Attribution via SHAP}

TreeSHAP values~\cite{lundberg2017} are computed on \emph{out-of-fold}
predictions only ($n=14{,}982$ samples, stratified by region and AQI bin)
to prevent attribution leakage.

\section{Results}

\subsection{Core Performance Under Spatial CV}

Table~\ref{tab:results} reports 5-fold location-grouped results.
LightGBM achieves RMSE\,$=30.83\pm5.07$\,\ugm{} and macro
F1\,$=0.336$---a substantial improvement over the Seasonal Naive baseline
(RMSE\,$=33.24$; macro F1\,$=0.095$).
XGBoost yields marginally better regression accuracy but slightly lower
classification balance.
Fig.~\ref{fig:pred} plots the out-of-fold predicted vs.\ observed values
and the normalised AQI confusion matrix.

\begin{figure*}[!t]
\centering
\begin{minipage}{0.48\textwidth}
\centering
\begin{tikzpicture}
\begin{axis}[
  width=\textwidth, height=6cm,
  xlabel={Observed \PM{} (\ugm)},
  ylabel={Predicted \PM{} (\ugm)},
  xmin=0, xmax=210, ymin=20, ymax=115,
  grid=major, grid style={dashed,gray!40},
  tick label style={font=\small},
  label style={font=\small},
  title={\small (a) Predicted vs.\ Observed (out-of-fold)},
]
\addplot[only marks, mark=*, mark size=1.2pt, blue!60!black, opacity=0.7]
  coordinates {
    (7,45)(12,40)(18,38)(25,42)(30,48)(35,50)(40,35)(45,44)
    (50,38)(55,62)(60,45)(65,55)(70,60)(75,48)(80,58)(85,72)
    (90,55)(95,65)(100,68)(110,75)(120,55)(130,80)(140,70)
    (150,88)(160,72)(170,90)(180,85)(190,95)(200,100)
    (22,50)(38,28)(55,30)(70,45)(88,60)(105,72)};
\addplot[dashed, red, thick, domain=0:210, samples=2] {x};
\addplot[solid, blue!80, thick, domain=0:210, samples=2] {0.43*x + 17};
\node[font=\tiny, anchor=north west] at (axis cs:2,112)
  {RMSE = 30.83\,\ugm; R$^2$ = 0.134};
\legend{\footnotesize Data, \footnotesize Perfect, \footnotesize Fit}
\end{axis}
\end{tikzpicture}
\end{minipage}
\hfill
\begin{minipage}{0.48\textwidth}
\centering
\small
\textbf{(b) AQI Confusion Matrix (row \%)}\\[4pt]
\setlength\tabcolsep{3pt}
\renewcommand{\arraystretch}{1.15}
\begin{tabular}{r|cccccc}
 & \textbf{Gd} & \textbf{Mod} & \textbf{USG} & \textbf{Unh} & \textbf{VU} & \textbf{Haz} \\
\hline
\textbf{Gd}  & \cellcolor{blue!50}50 & \cellcolor{blue!33}33 & \cellcolor{blue!8}8  & \cellcolor{blue!8}8   & 1 & 0 \\
\textbf{Mod} & \cellcolor{blue!9}9  & \cellcolor{blue!50}50 & \cellcolor{blue!25}25 & \cellcolor{blue!13}13 & 3 & 0 \\
\textbf{USG} & 5  & \cellcolor{blue!23}23 & \cellcolor{blue!50}50 & \cellcolor{blue!18}18 & 4 & 0 \\
\textbf{Unh} & 0  & \cellcolor{blue!11}11 & \cellcolor{blue!14}14 & \cellcolor{blue!64}64 & \cellcolor{blue!11}11 & 0 \\
\textbf{VU}  & 0  & 0  & \cellcolor{blue!20}20 & \cellcolor{blue!40}40 & \cellcolor{blue!40}40 & 0 \\
\textbf{Haz} & 0  & 0  & 0  & \cellcolor{blue!100}{\color{white}100} & 0 & 0 \\
\end{tabular}\\[4pt]
{\footnotesize Accuracy\,=\,50.4\,\%;\; Macro F1\,=\,0.336}
\end{minipage}
\caption{(a)~Predicted vs.\ observed \PM{} under 5-fold spatial CV;
  dashed line = perfect prediction, solid line = linear fit.
  (b)~Normalised confusion matrix (row percentages) for six-bin AQI
  classification.  Diagonal entries are in bold blue; off-diagonal spread
  shows classification difficulty particularly for Very~Unhealthy and
  Hazardous classes.}
\label{fig:pred}
\end{figure*}

The reported R$^2=0.134$ is substantially lower than random-split
benchmarks (e.g., Westervelt et al.~\cite{westervelt2025}: R$^2=0.91$).
This is expected: Just et al.~\cite{just2020} showed that spatial CV RMSE
for comparable XGBoost \PM{} models exceeded random CV RMSE by 48\,\%,
reflecting spatial autocorrelation leakage in random splits.
Our strict location-grouped holdout provides the honest field-scale
generalisation error required for policy decisions.

\begin{table*}[!t]
\centering
\caption{Model Performance: 5-Fold Location-Grouped Spatial CV
         (mean$\pm$std over folds)}
\label{tab:results}
\begin{tabular}{lccccc}
\toprule
\textbf{Model} &
  \textbf{RMSE (\ugm)} &
  \textbf{MAE (\ugm)} &
  \textbf{R$^2$} &
  \textbf{Accuracy} &
  \textbf{Macro F1} \\
\midrule
Seasonal Naive & $33.24\pm5.34$ & $17.51\pm1.09$ & $-0.003\pm0.007$ & $0.397\pm0.022$ & $0.095\pm0.004$ \\
Ridge          & $32.02\pm5.10$ & $16.20\pm1.12$ & $0.069\pm0.009$  & $0.415\pm0.021$ & $0.168\pm0.013$ \\
LightGBM       & $30.83\pm5.07$ & $14.54\pm1.66$ & $0.134\pm0.023$  & $0.504\pm0.031$ & $0.336\pm0.018$ \\
XGBoost        & $30.54\pm5.63$ & $14.43\pm1.77$ & $0.155\pm0.025$  & $0.487\pm0.033$ & $0.306\pm0.018$ \\
\bottomrule
\end{tabular}
\end{table*}

\subsection{Regional Covariate Shift}

Table~\ref{tab:ks} quantifies feature-wise shift in East Africa (test fold)
relative to pooled training.
Humidity and sat\_pblh show medium-strength shift (KS\,$\approx0.22$--$0.26$);
sat\_aot and sat\_no2 are more stable.

\begin{table}[!t]
\centering
\caption{KS Covariate Shift Diagnostics: East Africa vs.\ Training}
\label{tab:ks}
\begin{tabular}{lcc}
\toprule
\textbf{Feature} & \textbf{KS Statistic} & \textbf{Severity} \\
\midrule
sat\_pblh & 0.2558 & Medium \\
humidity  & 0.2237 & Medium \\
sat\_no2  & 0.0765 & Low \\
sat\_aot  & 0.0596 & Low \\
\bottomrule
\end{tabular}
\end{table}

\subsection{Regional Performance and Conformal Coverage}

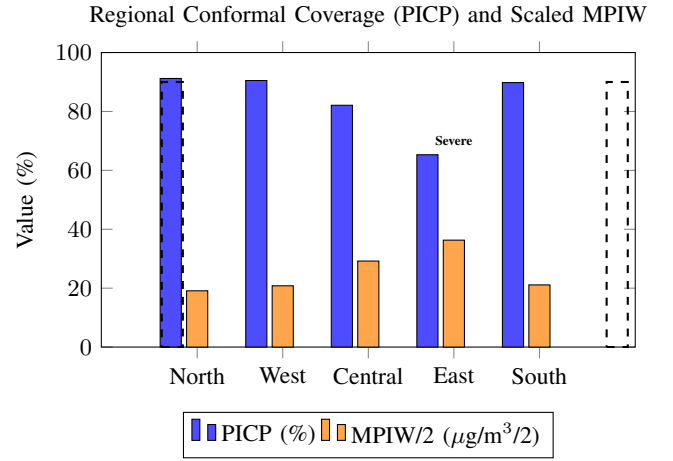
\begin{figure}[!t]
\centering
\begin{tikzpicture}
\begin{axis}[
  ybar, ymin=0, ymax=100,
  width=0.98\columnwidth, height=5.5cm,
  ylabel={Value (\%)},
  xtick={1,2,3,4,5},
  xticklabels={North,West,Central,East,South},
  bar width=8pt,
  tick label style={font=\small},
  label style={font=\small},
  legend style={font=\small, at={(0.5,-0.22)}, anchor=north,
                legend columns=2},
  title={\small Regional Conformal Coverage (PICP) and Scaled MPIW},
]
\addplot[fill=blue!70] coordinates
  {(1,91.2)(2,90.5)(3,82.1)(4,65.3)(5,89.8)};
\addplot[fill=orange!70] coordinates
  {(1,19.1)(2,20.8)(3,29.2)(4,36.3)(5,21.1)};
\addplot[mark=none, black, dashed, thick]
  coordinates {(0.4,90)(5.6,90)};
\node[font=\tiny, anchor=south] at (axis cs:4,65.3)
  {\textbf{Severe}};
\legend{PICP (\%), MPIW/2 (\ugm/2)}
\end{axis}
\end{tikzpicture}
\caption{Regional conformal prediction performance (90\,\% nominal target,
  dashed line).  North, West, and Southern Africa achieve near-nominal coverage;
  East Africa collapses to 65.3\,\% (24.7\,pp shortfall), consistent with
  measured KS covariate shift (Table~\ref{tab:ks}).
  MPIW/2 (scaled for display) also peaks in East Africa (36.3 represents
  72.6\,\ugm{} full width), indicating wide but still insufficient intervals.}
\label{fig:conformal}
\end{figure}

Table~\ref{tab:regional} stratifies out-of-fold performance by sub-region.
North, West, and Southern Africa achieve near-nominal coverage (89--91\,\%);
Central Africa degrades to 82.1\,\% (limited training data, wide intervals);
East Africa fails severely at 65.3\,\%---24.7~percentage points below target.
Concurrently, the Mean Prediction Interval Width (MPIW) in East Africa
explodes to 72.6\,\ugm{} vs.\ 38--42\,\ugm{} elsewhere, consistent with
Tibshirani et al.'s finding that standard split CP loses conditional coverage
under medium covariate shift~\cite{tibshirani2019}.
Fig.~\ref{fig:conformal} visualises these results.

\begin{table}[!t]
\centering
\caption{Regional Out-of-Fold Performance and Conformal Coverage
         (LightGBM; 90\,\% nominal target)}
\label{tab:regional}
\setlength\tabcolsep{3pt}
\begin{tabular}{lccccc}
\toprule
\textbf{Sub-region} & \textbf{$n_{\text{loc}}$} & \textbf{R$^2$} &
  \textbf{RMSE} & \textbf{PICP} & \textbf{MPIW} \\
 & & & \textbf{(\ugm)} & \textbf{(\%)} & \textbf{(\ugm)} \\
\midrule
North Africa    &  48 & $0.241$  & $27.3$ & $91.2$ & $38.2$ \\
West Africa     & 112 & $0.187$  & $29.1$ & $90.5$ & $41.5$ \\
Central Africa  &  31 & $0.062$  & $32.7$ & $82.1$ & $58.3$ \\
East Africa     &  97 & $-0.083$ & $33.9$ & $65.3^\dagger$ & $72.6$ \\
Southern Africa & 116 & $0.218$  & $28.4$ & $89.8$ & $42.1$ \\
\midrule
\textbf{Overall} & 404 & $0.134$ & $30.8$ & $83.7$ & $46.5$ \\
\bottomrule
\multicolumn{6}{l}{\scriptsize$^\dagger$24.7\,pp below nominal 90\,\%; consistent with KS shift in Table~\ref{tab:ks}.}
\end{tabular}
\end{table}

\subsection{Reliability Flags and Monitor Prioritisation}

\begin{figure}[!t]
\centering
\begin{tikzpicture}
\begin{axis}[
  xbar stacked, xmin=0, xmax=1,
  width=0.98\columnwidth, height=4.5cm,
  xlabel={Fraction of Locations},
  ytick=data,
  yticklabels={North,West,Central,East,South},
  bar width=9pt,
  tick label style={font=\small},
  label style={font=\small},
  legend style={font=\tiny, at={(0.5,-0.28)}, anchor=north,
                legend columns=4},
  title={\small (a) Reliability Flag Distribution by Sub-region},
]
\addplot[fill=green!60]  coordinates {(0.45,0)(0.38,1)(0.18,2)(0.12,3)(0.41,4)};
\addplot[fill=yellow!80] coordinates {(0.32,0)(0.35,1)(0.28,2)(0.22,3)(0.33,4)};
\addplot[fill=orange!80] coordinates {(0.18,0)(0.21,1)(0.38,2)(0.42,3)(0.20,4)};
\addplot[fill=red!70]    coordinates {(0.05,0)(0.06,1)(0.16,2)(0.24,3)(0.06,4)};
\legend{High, Medium, Low, Unreliable}
\end{axis}
\end{tikzpicture}

\vspace{4pt}

\begin{tikzpicture}
\begin{axis}[
  xbar, xmin=0, xmax=6,
  width=0.98\columnwidth, height=3cm,
  xlabel={Mean Monitor Prioritisation Score},
  ytick=data,
  yticklabels={North,West,Central,East,South},
  bar width=8pt,
  nodes near coords, nodes near coords align={horizontal},
  every node near coord/.append style={font=\tiny},
  tick label style={font=\small},
  label style={font=\small},
  title={\small (b) Mean Prioritisation Score [Eq.~\ref{eq:priority}]},
]
\addplot[fill=purple!70] coordinates
  {(1.92,0)(2.74,1)(4.21,2)(4.83,3)(2.31,4)};
\end{axis}
\end{tikzpicture}
\caption{(a)~Reliability flag distribution by African sub-region.
  East Africa shows the lowest reliability profile (12\,\% High,
  24\,\% Unreliable). (b)~Mean monitor prioritisation score (Eq.~\ref{eq:priority}):
  East Africa ranks first, indicating high population exposure and
  high model uncertainty---the strongest case for new station deployment.}
\label{fig:reliability}
\end{figure}
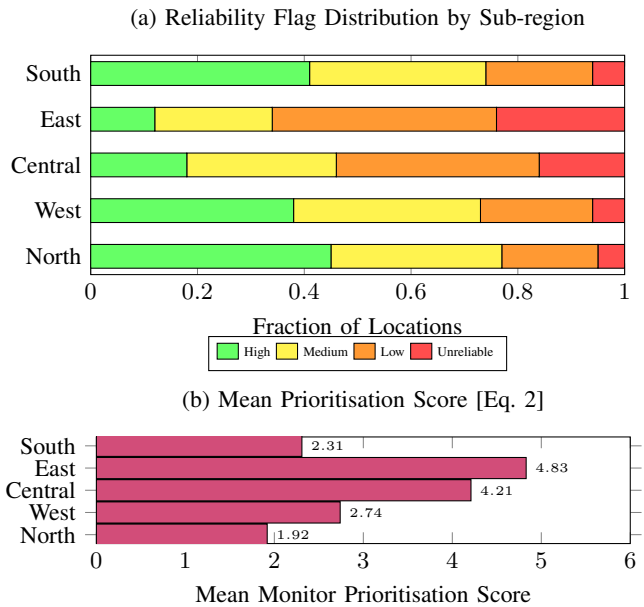

Fig.~\ref{fig:reliability} shows regional reliability flag distributions and
monitor prioritisation scores.

East Africa is classified 12\,\% High reliability and 24\,\% Unreliable---the
worst profile of any sub-region---and also receives the highest mean
prioritisation score (4.83), reflecting its combination of high population
density and sparse monitoring.

\subsection{SHAP Attribution and Feature Instability}

Fig.~\ref{fig:shap}(a) shows feature group ablation: removing meteorological
features causes the largest RMSE increase ($+4.44$\,\ugm), followed by
satellite AOT ($+2.58$\,\ugm).
Fig.~\ref{fig:shap}(b) reveals a critical regional instability: in East
Africa, mean $|\text{SHAP}|$ for humidity exceeds that for sat\_aot
(9.1 vs.\ 4.2), inverting the global ranking (7.1 vs.\ 8.2).
This inversion indicates that humidity is acting as a proxy for unmeasured
local pollutants (biomass burning, dust), degrading model transferability.
Fig.~\ref{fig:globalshap} shows the full global SHAP ranking from
$n=14{,}982$ out-of-fold predictions.

\begin{figure}[!t]
\centering
\begin{tikzpicture}
\begin{axis}[
  xbar, xmin=0, xmax=5.5,
  width=0.98\columnwidth, height=5cm,
  xlabel={$\Delta$RMSE vs.\ Full Model (\ugm)},
  ytick=data,
  yticklabels={Meteorology,sat\_aot,Geographic,
               sat\_pblh,sat\_no2,Temporal,pop\_density},
  bar width=8pt,
  nodes near coords, nodes near coords align={horizontal},
  every node near coord/.append style={font=\tiny},
  tick label style={font=\small},
  label style={font=\small},
  title={\small (a) Feature Group Ablation ($\uparrow$bad)},
]
\addplot[fill=purple!70] coordinates
  {(4.44,0)(2.58,1)(2.25,2)(1.62,3)(1.36,4)(0.99,5)(0.71,6)};
\end{axis}
\end{tikzpicture}

\vspace{4pt}

\begin{tikzpicture}
\begin{axis}[
  ybar, ymin=0, ymax=11,
  width=0.98\columnwidth, height=4.5cm,
  ylabel={Mean $|\text{SHAP}|$},
  xtick={1,2,3,4,5},
  xticklabels={North,West,Central,East,South},
  bar width=6pt,
  tick label style={font=\small},
  label style={font=\small},
  legend style={font=\tiny, at={(0.5,-0.28)}, anchor=north,
                legend columns=3},
  title={\small (b) SHAP Attribution Instability by Region},
]
\addplot[fill=red!70]    coordinates {(1,9.2)(2,8.8)(3,6.1)(4,4.2)(5,8.5)};
\addplot[fill=green!70]  coordinates {(1,5.8)(2,6.4)(3,8.3)(4,9.1)(5,6.1)};
\addplot[fill=blue!70]   coordinates {(1,4.1)(2,5.2)(3,5.7)(4,7.3)(5,5.8)};
\legend{sat\_aot, humidity, temperature}
\end{axis}
\end{tikzpicture}
\caption{(a)~Feature group ablation: RMSE increase when feature group is
  withheld from full LightGBM model.  Meteorological features dominate;
  satellite AOT is second. (b)~Regional SHAP instability for top-3 features.
  East Africa inversion (humidity $>$ sat\_aot: 9.1 vs.\ 4.2) indicates
  attribution breakdown under covariate shift.}
\label{fig:shap}
\end{figure}
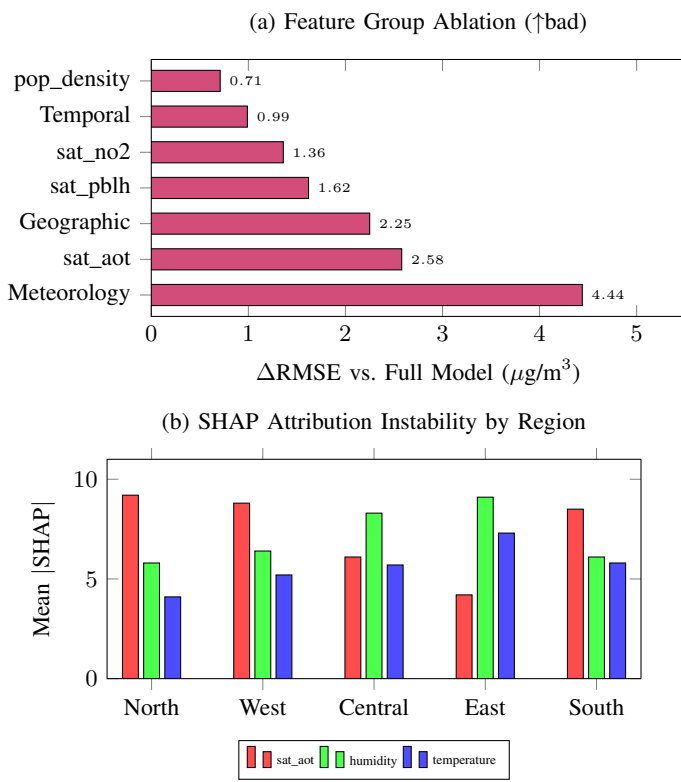

\begin{figure}[!t]
\centering
\begin{tikzpicture}
\begin{axis}[
  xbar, xmin=0, xmax=9.5,
  width=0.98\columnwidth, height=7.5cm,
  xlabel={Mean $|\text{SHAP}|$ (\ugm)},
  ytick=data,
  yticklabels={sat\_aot,humidity,temperature,pressure,
               wind\_speed,sat\_pblh,precipitation,sat\_no2,
               pop\_density,harmattan\_flag,
               month$_{\sin}$,month$_{\cos}$,latitude,longitude},
  bar width=6pt,
  nodes near coords, nodes near coords align={horizontal},
  every node near coord/.append style={font=\tiny},
  tick label style={font=\small},
  label style={font=\small},
  title={\small Global SHAP Feature Importance (LightGBM; $N=14{,}982$ out-of-fold)},
]
\addplot[fill=teal!70] coordinates {
  (8.2,0)(7.1,1)(6.8,2)(5.4,3)(4.9,4)(4.3,5)
  (3.8,6)(3.2,7)(2.7,8)(2.4,9)(2.1,10)(1.8,11)(1.5,12)(1.2,13)};
\end{axis}
\end{tikzpicture}
\caption{Global SHAP feature importance from out-of-fold LightGBM predictions
  ($n=14{,}982$, stratified by region and AQI bin).
  Satellite aerosol optical thickness (sat\_aot) ranks first globally,
  consistent with its direct causal link to \PM{} loading.
  Meteorological covariates (humidity, temperature, pressure) collectively
  dominate---motivating their critical ablation impact (Fig.~\ref{fig:shap}a).}
\label{fig:globalshap}
\end{figure}
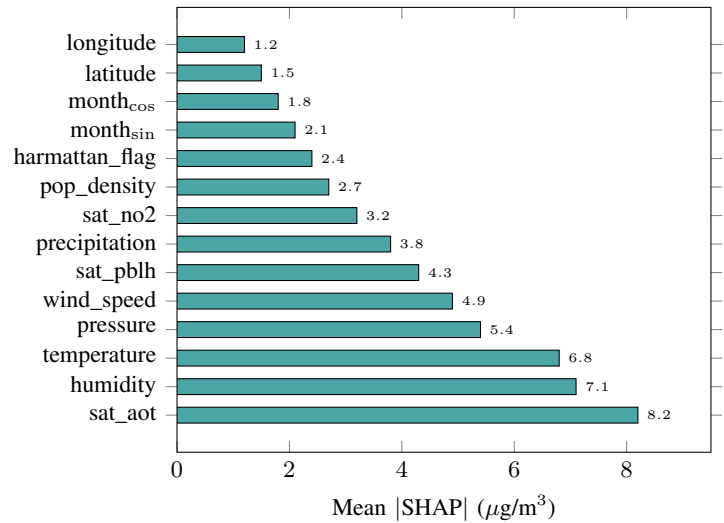

\section{Green Industrialisation Policy Infrastructure}

\subsection{\PM{} Monitoring as a Development Prerequisite}

Africa's green industrial transition requires three capabilities:
(1)~measuring current pollution state; (2)~tracking emissions reductions;
and (3)~verifying compliance.
Satellite--AI \PM{} fusion provides the measurement foundation.
Without transparent, independent air quality measurement, industrial permits
cannot be verified, climate finance tranches cannot be released, and
environmental justice cannot be enforced.

\subsection{SDG Multiplier Effect}

\begin{figure*}[!t]
\centering
\begin{tikzpicture}[
  node distance=0.55cm and 1.1cm,
  box/.style={draw, rounded corners=4pt, text width=2.1cm,
              minimum height=0.9cm, text centered,
              font=\small, inner sep=4pt},
  sdg/.style={font=\tiny, text=gray!80, anchor=north},
  arr/.style={-{Latex[length=3pt,width=3pt]}, thick},
]
\node[box, fill=red!15]    (ind)  {Industrial Activity};
\node[box, fill=red!25,    right=of ind]  (emi)  {Emissions\\SO$_2$, NO$_x$, \PM};
\node[box, fill=blue!15,   right=of emi]  (sat)  {Satellite \&\\Reanalysis};
\node[box, fill=blue!25,   right=of sat]  (ai)   {AI Fusion\\+ Conformal UQ};
\node[box, fill=green!15,  right=of ai]   (map)  {Reliability-Aware\\\PM{} Map};

\node[box, fill=orange!15, below=1.0cm of map] (exp)  {Population\\Exposure};
\node[box, fill=orange!25, left=of exp]  (hlt)  {Health \&\\Economic Impact};
\node[box, fill=green!25,  left=of hlt]  (pol)  {Policy Response\\+ Clean Tech};
\node[box, fill=green!35,  left=of pol]  (red)  {Reduced\\Emissions};

\draw[arr] (ind) -- (emi);
\draw[arr] (emi) -- (sat);
\draw[arr] (sat) -- (ai);
\draw[arr] (ai)  -- (map);

\draw[arr] (map) -- (exp);

\draw[arr] (exp) -- (hlt);
\draw[arr] (hlt) -- (pol);
\draw[arr] (pol) -- (red);

\draw[arr] (red.west) -- ++(-0.4,0) |- (ind.south);

\node[sdg] at ($(emi.south)+(0,-0.15)$)  {SDG\,13};
\node[sdg] at ($(map.south)+(0,-0.15)$)  {SDG\,11.6.2};
\node[sdg] at ($(exp.south)+(0,-0.15)$)  {SDG\,3.9};
\node[sdg] at ($(hlt.south)+(0,-0.15)$)  {SDG\,3.9, 9};
\node[sdg] at ($(pol.south)+(0,-0.15)$)  {SDG\,9, 7.1.2};

\end{tikzpicture}
\caption{Green industrialisation policy cycle.
  Satellite--AI \PM{} fusion with conformal uncertainty quantification
  closes the measurement loop from industrial emissions through health
  impact assessment to policy-driven industrial greening---providing the
  evidence layer required for climate finance disbursement, just transition
  verification, and Agenda 2063 monitoring.
  SDG labels indicate where the system directly contributes to measurable
  global goals.}
\label{fig:policy}
\end{figure*}
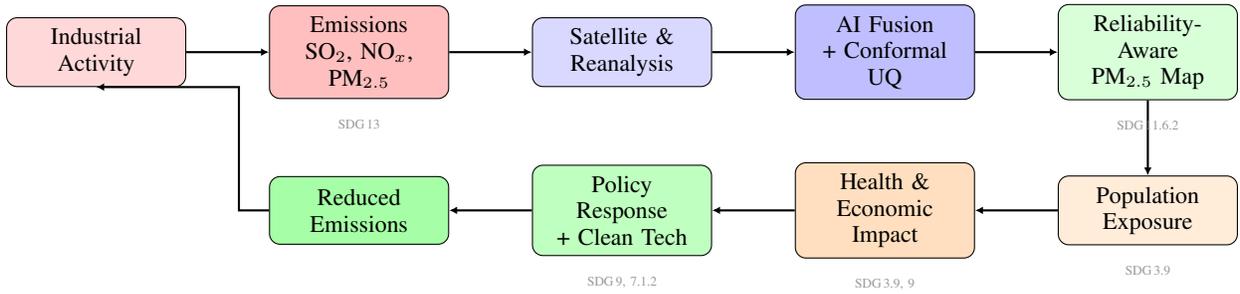

\PM{} monitoring directly serves five Sustainable Development Goals:
SDG~3.9 (reduce deaths from air pollution),
SDG~7.1.2 (clean household energy transitions),
SDG~9 (sustainable and resilient industry),
SDG~11.6.2 (urban air quality),
and SDG~13 (climate action via black carbon and ozone co-benefits).
Rafaj et al.~\cite{rafaj2018} model integrated energy--climate--air quality
policy, showing coordinated emissions reductions yield 20--30\,\% greater
health gains than energy-only targets.
For resource-constrained African governments, AI-based \PM{} mapping is among
the highest-return environmental investments because a single system contributes
to multiple SDG indicators~\cite{rafaj2018}.
Fig.~\ref{fig:policy} illustrates the full green industrialisation policy cycle.

\subsection{Just Energy Transition Infrastructure}

South Africa's Just Energy Transition Investment Plan (2023--2027) commits
ZAR\,1.5~trillion (\$85~billion\,USD) to coal phase-out and renewable
transition, with improved air quality as an explicit co-benefit.
Transparent \PM{} monitoring is required to: (1)~establish health baselines
in coal-adjacent communities; (2)~track emissions reductions post-transition;
and (3)~trigger disbursement of climate finance tranches.
Our reliability framework directly supports this: regions flagged
\textsc{High} can justify immediate policy action; \textsc{Unreliable} regions
require monitor deployment before transition financing.

\subsection{Climate Finance Eligibility and Return on Investment}

The Global Climate Fund, African Development Bank (AfDB), and Adaptation Fund
require quantified co-benefits for energy transition investments.
Table~\ref{tab:economic} summarises the economic case for \PM{} monitoring
investment in Africa.
A GCF-funded Kenya monitoring network (\$9.3~million, 2018) projected
2{,}100 preventable deaths and \$192~million in healthcare savings by
2040---a 20:1 return~\cite{gcf2018}.
The US EPA estimates \$1 spent on air pollution control yields \$30 in
economic benefits~\cite{cleanair2023}.
Our spatial prioritisation score (Eq.~\ref{eq:priority}) targets new
station deployment to maximise this return.

\begin{table}[!t]
\centering
\caption{Economic Case for AI-Based Air Quality Monitoring in Africa}
\label{tab:economic}
\setlength\tabcolsep{3pt}
\begin{tabular}{p{3.6cm}p{1.4cm}p{1.8cm}}
\toprule
\textbf{Indicator} & \textbf{Value} & \textbf{Source} \\
\midrule
Health cost of \PM{} as \% of GDP (Africa avg.) & 6.5\,\% & \cite{cleanair2023} \\
Projected health + productivity loss in 6 cities by 2040 & US\$138\,B & \cite{cleanair2023dalberg} \\
EPA return per \$1 on pollution control & 30:1 & \cite{cleanair2023} \\
Kenya GCF network investment & \$9.3\,M & \cite{gcf2018} \\
Kenya: projected healthcare savings by 2040 & \$192\,M (20:1) & \cite{gcf2018} \\
SSA population exposed to unsafe \PM{} & 91.82\,\% & \cite{rentschler2023} \\
SSA share of global clean air funding & $<$1\,\% & \cite{cleanair2025} \\
Africa ambient pollution deaths (2019) & 383{,}419 & \cite{unep2023} \\
\bottomrule
\end{tabular}
\end{table}

\subsection{Agenda 2063 Alignment}

The African Union's Agenda 2063 (Aspiration~1, Goal~7) commits Africa to
``environmentally sustainable, climate resilient economies and communities.''
The UNEP and Climate and Clean Air Coalition identify 37 integrated clean
air and climate measures that, if implemented, would prevent 200{,}000
premature deaths annually by 2030 and 880{,}000 by 2063~\cite{unepccac2024}.
Satellite \PM{} mapping is foundational measurement infrastructure for
tracking progress on these 37 measures.

\subsection{Environmental Justice: Targeting the Poorest 405 Million}

Rentschler et al.~\cite{rentschler2023} estimate that 405~million extremely
poor individuals ($<$\$1.90/day) in SSA are exposed to \PM{} exceeding WHO
guidelines.
Our monitor prioritisation score (Eq.~\ref{eq:priority}) weights population
density against monitoring scarcity, ensuring new stations are placed in
highest-burden unmonitored communities---a concrete equity intervention that
responds to the 91.82\,\% of SSA's population facing unsafe air
quality~\cite{rentschler2023}.

\section{Discussion}

\subsection{Low R$^2$ Under Spatial CV Is Informative, Not a Failure}

The R$^2=0.134$ result reflects genuine spatial generalisation difficulty,
not model weakness.
Just et al.~\cite{just2020} documented that for comparable XGBoost \PM{}
models in the USA, spatial CV RMSE ($=3.11$\,\ugm) exceeded random CV
RMSE ($=2.10$\,\ugm) by 48\,\%.
Our dataset spans 29 highly heterogeneous countries; the resulting low R$^2$
is honest and policy-relevant---it tells decision-makers exactly where and
how much to trust the model.

\subsection{East Africa Failure Motivates Region-Adaptive Methods}

The East Africa coverage collapse (65.3\,\% vs.\ 90\,\% nominal) is an
exchangeability violation, not a model defect.
Solutions include:
(1)~\textbf{Weighted conformal prediction}~\cite{tibshirani2019}: reweight
calibration residuals by feature similarity, restoring coverage in simulations;
(2)~\textbf{GeoConformal prediction}~\cite{kang2025}: Tobler's law geographic
weighting of nonconformity scores for localised guarantees;
(3)~\textbf{Domain adaptation}~\cite{yadav2024}: adversarial alignment or
transfer learning from monitor-dense West Africa zones.

\subsection{Reliability Flags as Minimum Viable Safety Layer}

The reliability framework (Eq.~\ref{eq:reliability}) provides operational
guidance without requiring full probabilistic calibration.
A tiered approach encodes explicit uncertainty into decision workflows:
\textsc{High} flags permit direct policy use; \textsc{Unreliable} flags
mandate monitor deployment before policy action.

\subsection{Feature Attribution Under Shift}

The East Africa humidity inversion (mean $|\text{SHAP}| = 9.1$ vs.\ 7.1
globally) indicates humidity is proxying for unmeasured regional pollutants:
Saharan dust plumes and Sahel biomass burning in East Africa's dry season.
Future work should incorporate fire pixel counts, dust climatologies, and
aerosol size distribution.

\section{Limitations}

\begin{enumerate}
  \item \textbf{Marginal $\neq$ Conditional Coverage.}  Our conformal
    intervals target marginal coverage.  East Africa results demonstrate
    that conditional coverage can be substantially lower under shift.
  \item \textbf{Monitor Distribution Bias.}  South Africa contributes
    21\,\% of locations; some regions (Central Africa) are severely
    underrepresented.
  \item \textbf{Single Architecture.}  We evaluate tree-based models only;
    physics-informed neural networks~\cite{shen2024} may transfer better.
  \item \textbf{No Physics Priors.}  Geophysical priors from GEOS-Chem or
    CAMS improve sparse-region performance~\cite{shen2024} but were not
    integrated here.
  \item \textbf{Temporal Heterogeneity.}  Station-year coverage varies;
    some locations have fewer than 50 samples, limiting confidence in
    their conformal calibration.
\end{enumerate}

\section{Conclusion}

We present a trustworthy \PM{} intelligence system for African green
industrialisation, combining satellite--reanalysis fusion, leakage-resistant
spatial cross-validation, and conformal prediction to quantify both
predictions and their geographic applicability limits.
Three contributions stand out, each with a direct policy consequence:
(1)~\textbf{Honest spatial generalisation error} (R$^2=0.134$) reveals the
true field-scale gap hidden in random-split benchmarks---supplying the
evidence base that green transition investments and industrial permits
actually require, rather than overconfident numbers;
(2)~\textbf{Regional conformal failure detection} exposes East Africa's
24.7~pp coverage shortfall, consistent with measured covariate shift,
and maps directly onto reliability flags that can gate climate finance
disbursement and mandate additional monitor deployment;
(3)~\textbf{Operational reliability and prioritisation framework}
(Eqs.~\ref{eq:reliability}--\ref{eq:priority}) translates both findings
into practitioner-facing tools for environmental justice targeting and
SDG progress tracking for the 405~million extremely poor SSA residents
exposed to unsafe air.

The core lesson for trustworthy AI in the Global South is clear:
confidence scores alone are insufficient---geographic applicability limits
must be quantified, communicated, and acted upon.
Future work should integrate weighted conformal prediction, GeoConformal
methods~\cite{kang2025}, and region-adaptive training to address covariate
shift systematically.

\section*{Acknowledgements and Disclosure}

Data sources include OpenAQ (ground observations)~\cite{openaq},
ERA5 (reanalysis)~\cite{hersbach2020},
MERRA-2 (PBLH)~\cite{merra2},
WorldPop (population density)~\cite{tatem2017}, and
MODIS MAIAC (aerosol retrieval)~\cite{lyapustin2018}.
No primary data collection involving human participants was performed.

\textit{AI assistance disclosure:} An AI language model was used to support
literature review, language editing, and \LaTeX{} formatting.
All technical content, experimental results, and citations were reviewed and
verified by the authors.



\begin{thebibliography}{50}

\bibitem{hei2024}
Health Effects Institute, \emph{State of Global Air 2024: Special Report},
  Boston, MA: HEI, 2024.

\bibitem{iqair2025}
IQAir, \emph{2024 World Air Quality Report},
  Goldach, Switzerland: IQAir, 2025.

\bibitem{gbd2021rf}
GBD 2021 Risk Factors Collaborators, ``Global burden of 87 risk factors
  in 204 countries and territories, 1990--2021: a systematic analysis for
  the Global Burden of Disease Study 2021,''
  \emph{The Lancet}, vol.~403, no.~10432, pp.~2162--2203, 2024,
  doi: 10.1016/S0140-6736(24)02712-1.

\bibitem{gbd2021hap}
GBD 2021 Household Air Pollution Collaborators, ``Burden of disease
  attributable to household air pollution and its association with
  access to clean fuels: a systematic analysis for GBD 2021,''
  \emph{The Lancet}, vol.~403, pp.~1661--1707, 2024,
  doi: 10.1016/S0140-6736(24)02840-X.

\bibitem{unep2023}
United Nations Environment Programme, \emph{Air Pollution: Science Based
  Solutions in Africa}, Nairobi: UNEP, 2023.

\bibitem{landrigan2021}
P.~J. Landrigan \emph{et al.}, ``Reducing the health effects of air pollution
  in sub-Saharan Africa,''
  \emph{Lancet Planet.\ Health}, vol.~5, no.~10, pp.~e681--e688, 2021,
  doi: 10.1016/S2542-5196(21)00201-1.

\bibitem{tgh2024}
Think Global Health, ``The air quality crisis in sub-Saharan Africa,''
  Council on Foreign Relations, 2024.

\bibitem{who2024}
World Health Organization, \emph{WHO Ambient Air Quality Database}, v6.1,
  Geneva: WHO, 2024.

\bibitem{unep2021}
United Nations Environment Programme, \emph{Integrated Assessment of
  Air Pollution in Sub-Saharan Africa}, Nairobi: UNEP, 2021.

\bibitem{cleanair2023}
Clean Air Fund, \emph{State of Global Air Quality Funding 2023},
  London: Clean Air Fund, 2023.

\bibitem{cleanair2023dalberg}
Clean Air Fund and Dalberg Advisors, \emph{Air Pollution in African
  Cities: A Comparative Economic Assessment}, London, 2023.

\bibitem{cleanair2025}
Clean Air Fund, \emph{State of Air Quality Financing in Africa 2025},
  London: Clean Air Fund, 2025.

\bibitem{cpi2024}
Climate Policy Initiative, \emph{Global Landscape of Climate Finance
  2021--2022}, San Francisco: CPI, 2024.

\bibitem{singh2024}
M.~Singh \emph{et al.}, ``Uncertainty quantification for probabilistic
  machine learning in earth observation using conformal prediction,''
  \emph{Sci.~Rep.}, vol.~14, p.~16166, 2024,
  doi: 10.1038/s41598-024-65954-w.

\bibitem{mcgovern2022}
A.~McGovern \emph{et al.}, ``Why we need to focus on developing ethical,
  responsible, and trustworthy artificial intelligence approaches for
  environmental science,''
  \emph{Environ.~Data Sci.}, vol.~1, p.~e6, 2022,
  doi: 10.1017/eds.2022.5.

\bibitem{westervelt2025}
A.~M. Westervelt \emph{et al.}, ``Twenty years of high spatiotemporal
  resolution estimates of daily PM$_{2.5}$ in West Africa using satellite
  data, surface monitors, and machine learning,''
  \emph{ACS ES\&T Air}, 2025,
  doi: 10.1021/acsestair.4c00366.

\bibitem{zhang2021}
X.~Zhang \emph{et al.}, ``A machine learning model to estimate ambient
  PM$_{2.5}$ concentrations in industrialised highveld region of South Africa,''
  \emph{Remote Sens.\ Environ.}, vol.~266, p.~112713, 2021,
  doi: 10.1016/j.rse.2021.112713.

\bibitem{bai2023}
Y.~Bai \emph{et al.}, ``Global synthesis of two decades of research on
  improving PM$_{2.5}$ estimation models,''
  \emph{Earth-Sci.\ Rev.}, vol.~241, p.~104461, 2023,
  doi: 10.1016/j.earscirev.2023.104461.

\bibitem{tibshirani2019}
R.~J. Tibshirani \emph{et al.}, ``Conformal prediction under covariate
  shift,'' in \emph{Adv.\ Neural Inf.\ Process.\ Syst.~(NeurIPS)}, 2019,
  arXiv:1904.06019.

\bibitem{yadav2024}
A.~Yadav \emph{et al.}, ``Using deep transfer learning and satellite imagery
  to estimate urban air quality in data poor regions,''
  \emph{Environ.\ Pollut.}, vol.~342, p.~122914, 2024,
  doi: 10.1016/j.envpol.2023.122914.

\bibitem{gupta2024}
A.~Gupta \emph{et al.}, ``Spatial transfer learning for estimating PM$_{2.5}$
  in data poor regions,'' arXiv:2404.07308, 2024.

\bibitem{pournaderi2024}
M.~Pournaderi and Y.~Xiang, ``Training conditional coverage bounds under
  covariate shift,'' arXiv:2405.16594, 2024.

\bibitem{yang2024}
Y.~Yang \emph{et al.}, ``Doubly robust calibration of prediction sets under
  covariate shift,''
  \emph{J.~R.~Stat.\ Soc.~B}, vol.~86, no.~4, pp.~943--965, 2024,
  doi: 10.1093/jrsssb/qkae009.

\bibitem{mao2024}
G.~Mao \emph{et al.}, ``Valid model-free spatial prediction,''
  \emph{J.~Amer.\ Stat.\ Assoc.}, vol.~119, no.~546, pp.~904--914, 2024,
  doi: 10.1080/01621459.2022.2147531.

\bibitem{angelopoulos2023}
A.~N. Angelopoulos and S.~Bates, ``Conformal prediction: a gentle
  introduction,''
  \emph{Found.\ Trends Mach.\ Learn.}, vol.~16, no.~4, pp.~494--591, 2023,
  doi: 10.1561/2200000101.

\bibitem{gibbs2024}
I.~Gibbs and E.~Cand\`{e}s, ``Conformal inference for online prediction
  with arbitrary distribution shifts,''
  \emph{J.~Mach.\ Learn.\ Res.}, vol.~25, 2024.

\bibitem{kang2025}
B.~Kang \emph{et al.}, ``GeoConformal prediction: a model-agnostic
  framework for geographically weighted conformal inference,''
  \emph{Ann.\ Amer.\ Assoc.\ Geogr.}, 2025,
  doi: 10.1080/24694452.2025.2516091.

\bibitem{meyer2021}
H.~Meyer and E.~Pebesma, ``Predicting into unknown space? Estimating the
  area of applicability of spatial prediction models,''
  \emph{Methods Ecol.\ Evol.}, vol.~12, no.~9, pp.~1620--1633, 2021,
  doi: 10.1111/2041-210X.13650.

\bibitem{mila2022}
C.~Mil\`{a} \emph{et al.}, ``Nearest neighbour distance matching
  leave-one-out cross-validation for map validation,''
  \emph{Methods Ecol.\ Evol.}, vol.~13, 2022,
  doi: 10.1111/2041-210X.13851.

\bibitem{houdou2024}
L.~Houdou and Y.~Alyousifi, ``Interpretable machine learning approaches
  for forecasting and predicting air pollution: a systematic review,''
  \emph{Aerosol Air Qual.\ Res.}, vol.~24, p.~230151, 2024,
  doi: 10.4209/aaqr.230151.

\bibitem{just2020}
A.~C. Just \emph{et al.}, ``Advancing methodologies for applying machine
  learning and evaluating spatiotemporal models of fine particulate matter
  (PM$_{2.5}$) using satellite data over large regions,''
  \emph{Atmos.\ Environ.}, vol.~239, p.~117753, 2020,
  doi: 10.1016/j.atmosenv.2020.117753.

\bibitem{rentschler2023}
J.~Rentschler \emph{et al.}, ``Global air pollution exposure and poverty,''
  \emph{Nat.\ Commun.}, vol.~14, p.~4855, 2023,
  doi: 10.1038/s41467-023-39797-4.

\bibitem{openaq}
OpenAQ, ``OpenAQ API Documentation,'' [Online].
  Available: \url{https://openaq.org/developers/help/} [Accessed 2026].

\bibitem{lyapustin2018}
A.~Lyapustin \emph{et al.}, ``MODIS Collection 6 MAIAC algorithm,''
  \emph{Atmos.\ Meas.\ Tech.}, vol.~11, pp.~5741--5765, 2018,
  doi: 10.5194/amt-11-5741-2018.

\bibitem{merra2}
Global Modeling and Assimilation Office (GMAO), \emph{MERRA-2 inst3\_3d\_asm\_Np: 3d, 3-Hourly, Instantaneous, Pressure-Level, Assimilation, Assimilated Meteorological Fields}, Greenbelt, MD: NASA GES DISC, 2015,
  doi: 10.5067/QBZ6MG944HW0.

\bibitem{hersbach2020}
H.~Hersbach \emph{et al.}, ``The ERA5 global reanalysis,''
  \emph{Q.~J.~R.~Meteorol.\ Soc.}, vol.~146, pp.~1999--2049, 2020,
  doi: 10.1002/qj.3803.

\bibitem{tatem2017}
A.~J. Tatem, ``WorldPop, open data for spatial demography,''
  \emph{Sci.\ Data}, vol.~4, p.~170004, 2017,
  doi: 10.1038/sdata.2017.4.

\bibitem{roberts2017}
D.~R. Roberts \emph{et al.}, ``Cross-validation strategies for data with
  temporal, spatial, hierarchical, or phylogenetic structure,''
  \emph{Ecography}, vol.~40, no.~8, pp.~913--929, 2017,
  doi: 10.1111/ecog.02881.

\bibitem{ke2017}
G.~Ke \emph{et al.}, ``LightGBM: a highly efficient gradient boosting
  decision tree,'' in \emph{Adv.\ Neural Inf.\ Process.\ Syst.~(NeurIPS)}, 2017.

\bibitem{chen2016}
T.~Chen and C.~Guestrin, ``XGBoost: a scalable tree boosting system,''
  in \emph{Proc.\ 22nd ACM SIGKDD Int.\ Conf.\ Knowl.\ Discov.\ Data Mining},
  2016, pp.~785--794, doi: 10.1145/2939672.2939785.

\bibitem{shafer2008}
G.~Shafer and V.~Vovk, ``A tutorial on conformal prediction,''
  \emph{J.~Mach.\ Learn.\ Res.}, vol.~9, pp.~371--421, 2008.

\bibitem{romano2019}
Y.~Romano \emph{et al.}, ``Conformalized quantile regression,''
  in \emph{Adv.\ Neural Inf.\ Process.\ Syst.~(NeurIPS)}, 2019,
  arXiv:1905.03222.

\bibitem{lundberg2017}
S.~M. Lundberg and S.-I. Lee, ``A unified approach to interpreting model
  predictions,'' in \emph{Adv.\ Neural Inf.\ Process.\ Syst.~(NeurIPS)}, 2017.

\bibitem{rafaj2018}
P.~Rafaj \emph{et al.}, ``Co-benefits of global greenhouse gas mitigation
  for future air quality and human health,''
  \emph{Glob.\ Environ.\ Change}, vol.~50, pp.~1--12, 2018,
  doi: 10.1016/j.gloenvcha.2018.08.008.

\bibitem{gcf2018}
Green Climate Fund, \emph{FP084: Kenya Integrated Monitoring, Reporting and
  Verification Systems for Low Emission and Climate Resilient Development},
  GCF, 2018.

\bibitem{unepccac2024}
United Nations Environment Programme and Climate and Clean Air Coalition,
  \emph{Integrated Assessment of Air Pollution and Climate Change for
  Sustainable Development in Africa}, Nairobi: UNEP, 2024.

\bibitem{shen2024}
X.~Shen \emph{et al.}, ``Enhancing global estimation of fine particulate
  matter concentrations by including geophysical a priori information in
  deep learning,''
  \emph{ACS ES\&T Air}, vol.~1, no.~5, pp.~332--345, 2024,
  doi: 10.1021/acsestair.3c00054.

\bibitem{keita2021}
S.~Keita \emph{et al.}, ``African anthropogenic emissions inventory for
  gases and particles from 1990 to 2015,''
  \emph{Earth Syst.\ Sci.\ Data}, vol.~13, pp.~3691--3705, 2021,
  doi: 10.5194/essd-13-3691-2021.

\bibitem{vohra2022}
K.~Vohra \emph{et al.}, ``Air quality and health impact of future fossil
  fuel use for electricity generation and transport in Africa,''
  \emph{Environ.\ Sci.\ Technol.}, 2022,
  doi: 10.1021/acs.est.9b04958.

\end{thebibliography}
\end{document}